%% file: 0-iswc-resource-diso-main.tex
\documentclass[runningheads]{llncs}
\usepackage[T1]{fontenc}
\usepackage[utf8]{inputenc}
\usepackage{times}
\usepackage{hyperref}
\usepackage{amsmath}
\usepackage{amssymb}
\usepackage{subcaption}
\usepackage{graphicx}
\usepackage{algorithm}
\usepackage{algpseudocode}
\usepackage{booktabs} 
\usepackage{enumitem}
\usepackage{todonotes}
\usepackage{threeparttable}
\usepackage{multirow}
\usepackage{pgf}
\usepackage{pgfplots}

\usepackage{xspace}
\newcommand{\ie}{\textit{i}.\textit{e}.,\xspace}
\newcommand{\eg}{\textit{e}.\textit{g}.,\xspace}

\newcommand{\DNS}{\textit{defence and national security}\xspace}
\newcommand{\DNSU}{\textit{Defence and national security}\xspace}

\input{macros}

\begin{document}

\def\thefootnote{\dag}\footnotetext{These authors contributed equally to this work.}\def\thefootnote{\arabic{footnote}}

%
\title{Improving Interoperability among Defence and National Security Ontologies: Analysis and Evaluation Tasks}
%
\titlerunning{Improving Interoperability among Defence and National Security Ontologies}
%
\author{
Jonathon Dilworth\inst{\dag,1}\orcidID{0009-0009-0260-0492} \and
Pedro Giesteira Cotovio \inst{\dag,1,2}\orcidID{0000-0001-6724-899X} \and
David Herron\inst{\dag,1}\orcidID{0009-0008-2736-6789} \and  
Paul Cripps\inst{3} \and
Nigel Dewdney\inst{4}
\and
Catia Pesquita\inst{2}\orcidID{0000-0002-1847-9393} 
\and
Ernesto Jiménez-Ruiz\thanks{Corresponding author.}\inst{1}\orcidID{0000-0002-9083-4599}}
\authorrunning{J. Dilworth et al.}
%
\institute{
City St George’s, University of London, London, UK\\
\email{ernesto.jimenez-ruiz@citystgeorges.ac.uk}\\ \and
LASIGE, Faculdade de Ciências, Universidade de Lisboa, Portugal \and
Defence Science and Technology Laboratory, UK \and
The Alan Turing Institute, UK
}

\maketitle              
\begin{abstract}
The use of ontologies and knowledge graphs is becoming increasingly widespread in the \DNS domain. Numerous ontologies have been developed through initiatives led by academia, industry, and government. Achieving interoperability across diverse \DNS ontologies remains a major challenge due to the domain’s breadth and specialisation. In this work, we analyse and document over 60 publicly available ontologies and introduce a new track for the Ontology Alignment Evaluation Initiative (OAEI). This track comprises eight matching tasks, consensus alignments and manually-curated (silver-standard) mappings. The consensus alignments are derived by aggregating the outputs of several state-of-the-art ontology alignment systems. The silver-standard is obtained from the manual validation of the consensus alignment together with a subset of the unique mappings (\ie mappings suggested by only one system).

\medskip

\textbf{Resource Type:} Dataset and benchmark.

\textbf{DISO Repository:} 
{\scriptsize
\url{https://github.com/city-artificial-intelligence/diso}}

\textbf{OAEI Track:} {\scriptsize
\url{https://city-artificial-intelligence.github.io/diso-oaei/}}

\textbf{Codes:} {\scriptsize
\url{https://github.com/city-artificial-intelligence/DISO-mappings}}

\textbf{Permanent repository:} {\scriptsize{\url{https://doi.org/10.5281/zenodo.20059506}}}

\textbf{License:} MIT License


\end{abstract}

\setcounter{footnote}{0} 

\input{1-introduction}

\input{2-Background}

\input{3-diso-oaei}

\input{4-conclusions}


\subsubsection*{Acknowledgements.}

This research was supported by Turing Innovations Limited and The Alan Turing Institute’s Defence and Security Programme via the project \href{https://ernestojimenezruiz.github.io/projects/guard/}{GUARD}. It was also supported by FCT through the fellowship  \url{https://doi.org/10.54499/2022.10557.BD} (Pedro Cotovio), the LASIGE Research Unit, ref. UID/00408/2025 and the CancerScan project, which received funding from the European Union’s Horizon Europe Research and Innovation Action (EIC Pathfinder Open) under grant agreement No. 101186829.
Views and opinions expressed are, however, those of the author(s) only and do not necessarily reflect those of The Alan Turing Institute, the European Union or the European Innovation Council and SMEs Executive Agency. Neither the European Union nor the granting authority can be held responsible for them.

\subsubsection*{Declaration of use of Generative AI.}
AI tools were used solely for grammar correction and minor language edits. They did not contribute to content creation, idea development, or substantive rewriting. The research design, experiments, analysis, and writing were carried out entirely by the authors.

\bibliographystyle{splncs04}
\bibliography{references-new,disobib}

\end{document}

%% file: macros.tex
\newcommand{\On}{\ensuremath{\mathcal{O}}\xspace}
\newcommand{\M}{\ensuremath{\mathcal{M}}}

\newcommand{\mapping}[4]{\langle #1,\allowbreak #2,\allowbreak #3,\allowbreak #4
\rangle}

\newcommand{\myset}[1]{ \{#1\} }
\newcommand{\MS}{\M^{S}\xspace}

\newcommand{\logicalalgo}[1]{\ifmmode \text{ \textbf{#1} } \else \textbf{#1} \fi}

\definecolor{drot}{rgb}{0.8,0.0,0.0}
\definecolor{dgruen}{rgb}{0.0,0.375,0.0}
\definecolor{dblue}{rgb}{0.0,0.0,0.375}
\definecolor{dbraun}{rgb}{0.3,0.2,0.1}

%% file: 1-introduction.tex
\section{Motivation}
\label{sec:intro}

\DNSU is one of the main UKRI\footnote{UK Research and Innovation (UKRI) funding body.} priorities, where AI techniques can play a crucial role in preventing and mitigating threats.
Indeed, the work presented in this paper is conducted within the scope of the GUARD project,\footnote{GUARD: \url{https://ernestojimenezruiz.github.io/projects/guard/}} which focuses on \textit{Ensuring Interoperable and Trustworthy Knowledge Graphs for Defence and National Security AI}. GUARD has been funded by The Turing Defence \& Security Grand Challenge, a collaborative effort towards meeting the strategic requirements of UK Government agencies for \emph{systematic sensemaking at scale and pace}.

The development of hybrid learning and reasoning systems in general, and Neurosymbolic (NeSy) AI systems \cite{DBLP:journals/air/GarcezL23} in particular, is gaining increasing attention to overcome the challenges of purely data-driven models (\eg LLMs - Large Language Models \cite{10.1145/3744746}) with respect to fairness, privacy, correctness, data and energy efficiency, identifiability, and eXplainability (XAI). 
In a domain like \DNS, the design and development of robust, reliable and semantically sound AI models is paramount to support the notion of systematic sensemaking.
Knowledge Graphs (KGs)~\cite{kgs2022}  play a key role in orchestrating diverse and heterogeneous data sources, while providing a semantic and mathematical (\ie logic-based) representation of the domain. KGs are becoming essential components of NeSy systems~\cite{city32688} to \textit{(i)} increase the coverage, validity and quality of the available data, \textit{(ii)} impose constraints during the learning process, \textit{(iii)} reason about what has been learned (\ie validity of the predictions), and, ultimately, \textit{(iv)} enable well-informed decision-making.

One of the main challenges when working with KGs is the definition of an ontology to model the domain. The domain for a \DNS application can be, however, quite broad as it may require the interplay of several subdomains to cover, among others,  threats (\eg physical and cyber), relevant actors (\eg government, terrorist groups, intelligence services), operations (\eg military, emergency response), infrastructure (\eg airport, bases), policies (\eg defence strategy, international law), and geolocations.

There are several contributions from academia, industry, and government to create ontologies and knowledge graphs for security and defence. Prominent examples are the general-purpose models IES\footnote{\url{https://github.com/IES-Org/ont-ies}} and DoDAF.\footnote{\url{https://dodcio.defense.gov/Library/DoD-Architecture-Framework/}} IES (Information Exchange Standard) has been developed by the UK Government; while the DoDAF Formal Ontology has been implemented by the Department of War/Defense Chief Information Officer and serves as a NATO/multi-national model for defence.
The IES models embrace semantic web technologies to represent general domain concepts such as locations, legal events, states, and measures. 
MITRE, funded by the National Security Agency, has also launched a family of cybersecurity ontologies to formalise offensive and defensive techniques \cite{Kaloroumakis-D3FEND-2021}.\footnote{\url{https://d3fend.mitre.org/}} 
The Department of Homeland Security has also designed a core Ontology for the maritime domain.\footnote{\url{https://bit.ly/ontology-maritime-domain}}
Efforts from academia include ontologies for military (\eg \cite{onto-cif,10.1007/978-981-99-3300-6_17,militart-jowo,DBLP:conf/stids/MorosoffRBFS15}), national security (\eg \cite{DBLP:conf/stids/CostaCPASS14,DBLP:conf/stids/MundieRDPMC14,DBLP:journals/ijcip/ChenLZHQ23}), and cybersecurity (\eg \cite{DBLP:conf/stids/ObrstCM12,DBLP:conf/stids/OltramariCWM14,DBLP:journals/cybersec/WangZLS21,DBLP:conf/IEEEares/ValeroMSNLLVMPP24}).
There are also prominent examples of collaborations among government, academia and industry. For example, the Intelligence Community (IC) Ontology Working Group (DIOWG), from the US Department of Defense, is leading on the development and implementation of a National Security Ontology Foundry (NSOF), which will rely on the Basic Formal Ontology (BFO),\footnote{\url{https://basic-formal-ontology.org/}} and the Common Core Ontology (CCO)\footnote{\url{https://github.com/CommonCoreOntology/CommonCoreOntologies}} as their baseline standards.\footnote{\url{https://bit.ly/us-dod-ontology-news}}
BFO is an academia-driven upper-level ontology, widely adopted in life sciences ontologies (\eg OBO Foundry\footnote{\url{https://obofoundry.org/}}), but with an increasing presence in government-related ontologies.
NATO are also actively investigating the application of ontologies in the Command and Control space. As part of their work on the Data Centric Reference Architecture for the Alliance (DCRA),\footnote{NATO. Data Centric Reference Architecture for the Alliance (2025): \url{https://nhqc3s.hq.nato.int/apps/DCRA_Report/}} they have outlined the need for a domain upper ontology for defence and security (CXCSRM) and ontology management services, with objectives closely aligned to those of the NSOF.

The interoperability among ontologies at the upper- and mid-level has been facilitated by efforts like IES, BFO and CCO. However, although IES and BFO are, in principle, compatible, a complete and formal alignment still needs to be established~\cite{DBLP:conf/jowo/BaileyBBCCCDHLM25}.
Furthermore, achieving interoperability across application ontologies may be particularly challenging in domains that require reconciling both the wide-ranging scope of concepts and the highly specialised representations in the subfields. 
\DNSU is an example of such domains. 
An ontology foundry for \DNS, as the one proposed by the DIOWG,\footnote{Recently rebranded as NSOWG (National Security Ontology Working Group).} should ensure
that its application ontologies achieve interoperability, comprehensive coverage, and high quality. Without these guarantees, the deployment of ontologies and KGs 
risks limiting their effectiveness in supporting downstream \DNS tasks.

In this paper, we focus on the analysis of the interoperability among \DNS ontologies. Our contributions are summarised as follows. \textit{(i)} We have collected and documented 60+ public ontologies relevant to the \DNS domain. \textit{(ii)} We have analysed their intersection by performing ontology alignment over 1,653 ontology pairs. \textit{(iii)} We have designed a new OAEI track including 8 matching tasks where we have compared the outcomes of several state-of-the-art alignment systems, created a consensus-based alignment, and manually verified a silver-standard reference alignment. The silver-standard has been generated by manually curating the consensus alignment and a subset of the unique system-generated mappings (\ie those suggested by only one system). 
The rest of the paper is organised as follows. Section \ref{sec:background} introduces the Defence, Intelligence and Security Ontologies (DISO) network, and the motivation for gathering and integrating these ontologies. The new DISO-OAEI evaluation track is described in Section \ref{sec:diso-oaei}, together with a comparison among the state-of-the-art systems.  Section \ref{sec:ecosystem} summarises the created resources. Finally, future work directions and conclusions are given in Section \ref{sec:conclusions}.

%% file: 2-Background.tex
\input{table-diso}

\section{Defence, Intelligence and Security Ontologies (DISO)}
\label{sec:background}

DISO (Defence, Intelligence and Security Ontologies)\footnote{\url{https://github.com/city-artificial-intelligence/diso/}} is a collection of 60+ publicly available Web Ontology Language (OWL) \cite{owl2008} ontologies related to the \DNS domain.
The ontologies in DISO were identified and obtained during an ontology search exercise undertaken primarily during Nov/Dec of 2025, drawing on the academic literature (including surveys of cyber-security ontologies, \eg \cite{DBLP:journals/corr/abs-2510-16610,Rivadeneira2021,DBLP:journals/kais/Sikos23}), government and standards-body materials, and public ontology registries and repositories.
An ontology was included in the collection if \textit{(i)} it is publicly available, \textit{(ii)} it is provided in (or has an available serialisation in) OWL, and \textit{(iii)} it is relevant to the \DNS domain, either directly (\ie developed by or for defence, intelligence and security organisations) or indirectly, through subdomains, such as `situation awareness' or `smart environments', that underpin \DNS applications (as discussed below). 
Relevant ontologies distributed without a permissive (or confirmed) licence are not redistributed, but linked from their original locations. The provenance of each ontology (upstream source and licence) is recorded in the DISO repository, and future extensions of the collection are expected to follow the same criteria through a documented contribution protocol.

\begin{figure}[t]
    \center
    \includegraphics[width=0.9\textwidth]{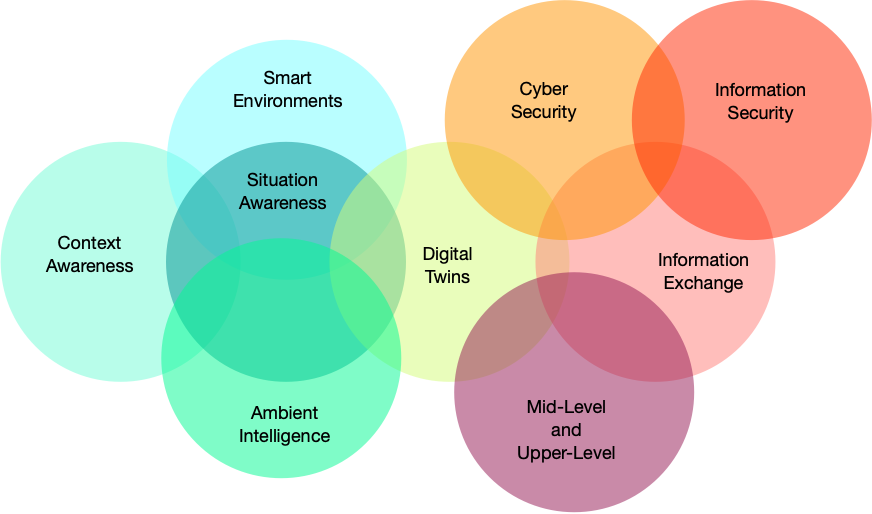}
    \caption{A view of the conceptual overlaps that exist between some of the clusters (or subdomains) of the ontologies within the DISO (Defence, Intelligence and Security Ontology) collection.} 
    \label{fig:diso-cluster-overlaps}
\end{figure}

The DISO collection has been clustered into \DNS categories (or subdomains).
Table \ref{table:diso-ontology-counts-per-cluster} shows the 11 clusters (or subdomains) of the DISO collection in alphabetic order along with the number of ontologies assigned to each cluster.
Three factors make it difficult to provide precise counts of the number of ontologies in the DISO collection and within its component clusters.
First, many of the ontologies obtained for DISO are physically represented as networks of multiple component ontology files.
SOUPA \cite{DBLP:conf/mobiquitous/ChenPFJ04}, for example, an ontology for ubiquitous and pervasive computing whose development was partly funded by DARPA, and which is assigned to the DISO cluster `situation awareness', is presented as a network of 18 component ontology files.
For DISO ontology counting purposes, however, we count SOUPA as a single ontology.
Hence, in this respect, the ontology counts reported in Table \ref{table:diso-ontology-counts-per-cluster} significantly understate the number of physically distinct ontology files present within the DISO repository.
Second, to make it easy for researchers to reuse such networked DISO ontologies, we have frequently merged such networks into single, all-inclusive ontology files.
Third, some ontologies within DISO fit naturally into more than one DISO cluster.
For example, the SOUPA ontology is regarded as being a member of both the `situation awareness' cluster and the `context awareness' cluster.
Figure \ref{fig:diso-cluster-overlaps} illustrates the conceptual overlap that exists between the clusters (subdomains). 

 Regarding the FAIRness~\cite{garijo2020fairvocabularies} of DISO , we have created metadata in RDF, extending the VoID and Dublin Core Terms vocabularies, to enhance the exploration of the DISO clusters and ontologies. The metadata includes links to the ontologies (remote and local versions), as well as their license. We have also added the repository to the LoD cloud.\footnote{\url{https://lod-cloud.net/dataset/DISO}} to improve its findability.
 The DISO clusters and ontologies are also documented in our GitHub repository and summarised as follows:

\smallskip
\noindent
\textit{Situation awareness.}
Situation awareness~\cite{DBLP:journals/hf/Endsley95,DBLP:journals/hf/Endsley2001} has to do with perceiving and understanding the elements of a dynamic environment, and the relations between them, whilst considering both space and time, and the prediction of the environment's future state(s).
The generality of the notion of situation awareness is reflected in the number of other DISO clusters with which it has close conceptual associations.

\smallskip
\noindent
\textit{Context awareness.}
`Context awareness', for instance, has primarily to do with individuals and their immediate surroundings: their nearby environmental context.
The common thread (and motivation) shared by the ontologies in this cluster is the idea of pervasive computing, where the objective is for mobile application services to be able to adapt their behaviour dynamically based on an application user's perceived current environmental context.
Thus, the notion of context awareness can be regarded as a specialised form of situation awareness, where the situation of interest is the surroundings of a mobile, software application user.

\smallskip
\noindent
\textit{Smart environments.}
It is similarly reasonable to regard smart environments (whether they be homes, or buildings, or cities) as being specialised forms of situations: built environment situations of various size and scale.
The notion of `smart' links to the notion of `awareness' because the ontologies of the DISO `smart environments' cluster generally presume some amount of sensor-based monitoring of the environment in question.
For example, the DISO `smart homes' ontology ThinkHome \cite{DBLP:journals/ejes/ReinischKIK11} has an emphasis on the monitoring and control of home energy use.

\smallskip
\noindent
\textit{Ambient intelligence.}
Ambient intelligence \cite{DBLP:journals/comsis/AugustoM07} models devices that seamlessly interconnect and collaborate with each other as well as with human users.
Like context awareness, ambient intelligence supports the idea of pervasive computing, but with more attention to supporting ideas like the Internet of Things (IoT). 
BOnSAI \cite{DBLP:conf/wims/StavropoulosVVB12} is a smart building ontology for ambient intelligence, and is assigned to both the `smart buildings' and `ambient intelligence' DISO clusters.

\smallskip
\noindent
\textit{Digital twins.}
Ontologies are well-suited to supporting the realisation of digital twins~\cite{Maryasin2019}, and their use in connection with digital twins has been examined in a recent survey \cite{DBLP:journals/fgcs/KarabulutPGD24}.
The notions of situation awareness, smart environments and ambient intelligence relate strongly to the notion of digital twins.
Indeed, what DISO refers to as smart environments (smart homes, buildings and cities) are prime examples of the types of physical things that the UK's National Digital Twin Programme (NDTP)\footnote{\url{https://ndtp.co.uk}} envisages having digital twins.
SAREF (a Smart Applications REFerence ontology)\footnote{\url{https://saref.etsi.org/}}, created by the European Telecommunications Standards Institute (ETSI), 
is one of the main ontologies discussed in connection with creating digital twins for buildings in \cite{Maryasin2019}.

\smallskip
\noindent
\textit{Information security.}
Information security has to do with protecting information in all its forms, whether physical or digital, tangible (\eg paperwork) or intangible (\eg knowledge), from things like unauthorised access or inappropriate use, modification, disclosure or corruption. 
DISO contains five ontologies covering this domain.
One of these, MDISOnt \cite{DBLP:journals/kais/MeriahRK25}, a multi-dimensional information security ontology, decomposes information security into several perspectives using dimensional views and modules.

\smallskip
\noindent
\textit{Cyber-security.}
Cybersecurity is widely understood as a specialised form of information security that focuses on the digital domain and the protection of assets against digital threats.
Three recent surveys of cyber-security ontologies \cite{DBLP:journals/corr/abs-2510-16610,Rivadeneira2021,DBLP:journals/kais/Sikos23} attest to the fact that the cyber-security domain appears to appeal to ontologists.
The DISO collection includes 15 such ontologies that are widely recognised in the literature.
A prominent example is the MITRE Corporation D3FEND cyber-security ontology \cite{Kaloroumakis-D3FEND-2021}.

\smallskip
\noindent
\textit{Information exchange.}
Ontologies in this cluster seek to standardise data representations so as to facilitate the exchange (and aggregation) of data across diverse domains, sectors, organisations and systems.
The DISO collection contains two authoritative information exchange ontologies.
One, IES (Information Exchange Standard)\footnote{\url{https://informationexchangestandard.org/}}, is a standard for information exchange developed within the UK Government.
DISO contains the IES version released in November 2025.
The IES plays a critical role in enabling the UK's vision of a National Digital Twin (NDT).
The other information exchange ontology in DISO is JC3IEDM,
standardised by NATO and jointly developed under the Multilateral Interoperability Programme (MIP) for the exchange of consultation, command and control (C3) information.

\smallskip
\noindent
\textit{Mid-level and Upper-level.}
In addition to domain/application level ontologies, DISO also contains clusters for mid-level and upper-level ontologies as well.
Upper-level ontologies seek to describe general concepts to embrace and orchestrate domain-level ontologies. 
Mid-level ontologies specialise upper-level ontologies for particular (vertical) domains, and thereby play a fundamental bridging role between (general) upper-level ontologies and (specialised) domain-level ontologies.
The criticality of mid-level and upper-level ontologies for enabling integration of diverse, heterogeneous knowledge sources is evidenced by the next version of IES, which is being modularised with `ies-top' and `ies-core' to provide the common foundations for IES domain and sub-domain level ontologies.

\subsection{Towards a network of DISO ontologies}

Ontology alignment \cite{ombook2013}  is the process of finding correspondences or an \emph{alignment} $\M$ among the entities (ontology classes, properties or instances) of two or more ontologies. 
A \emph{mapping} involving two entities is typically represented as a 4-tuple $\mapping{e_1}{e_2}{r}{c}$ where $e_1$ and $e_2$ are entities of the ontologies $\On_1$ and $\On_2$, 
respectively, $r$ is a semantic relation, typically one of $\myset{\sqsubseteq, \sqsupseteq, \equiv}$, and $c$ is a confidence value (usually a number between $0$ and $1$).
An ontology \emph{alignment system} is a program
that, given as input two ontologies, 
generates an ontology alignment $\MS$ between them.

\input{table-logmap-diso-all}

We conducted a bulk ontology alignment exercise using the ontology alignment system LogMap~\cite{logmap2011,logmap2012}  over the DISO ontologies, resulting in the (potential) generation of alignments for 1,653 ontology pairs.
Table \ref{table:logmap-alignment-size-analysis} compares the distributions of alignment sizes with respect to two different sets of DISO ontologies: \textit{(i)} the full set of 1,653 DISO ontology pairs, and \textit{(ii)} the subset of 105 DISO cyber-security ontology pairs derived from the 15 cyber-security ontologies. We observe that LogMap identifies an intersection between at least 647 ontology pairs (66 about cyber-security), indicating that the DISO ontology network has an important number of connections (\ie ontology mappings) across its ontologies as we advanced in Figure \ref{fig:diso-cluster-overlaps}.
The analysis of the bulk ontology alignment served as the basis to select a subset of ontology pairs for a new ontology alignment track, as described in the next section. Ultimately, the goal is to develop an integrated network of DISO ontologies that contributes to addressing the interoperability challenge in the \DNS domain.

%% file: table-diso.tex
\begin{table}[t]
  \centering
  \caption{A view of the conceptual structure of the DISO collection of defence, intelligence and security ontologies. The clusters (or subdomains) of the DISO collection are listed together with the number of ontologies assigned to each.}
  \label{table:diso-ontology-counts-per-cluster}
  \begin{tabular}{lrr}
    \toprule 
    DISO  Cluster/Subcluster Name~~~ & 
    ~~~~Cluster Size & 
    ~~~~Subcluster Size \\
    \midrule
    agentic  & 1  \\
    ambient intelligence & 1 \\
    context awareness  & 3 \\
    cyber-security & 15 \\
    digital twins & 1 \\
    information exchange & 2 \\
    information security & 5 \\
    mid-level &  12 \\
    risk management & 1 \\
    robotics & 2 \\
    situation awareness &  4 \\
    smart environments & 8 \\
    >> smart buildings &  & 4 \\
    >> smart cities &  & 2 \\
    >> smart homes &  & 2 \\
    upper-level & 8 \\
    \hline 
    totals & 63 \\
    \bottomrule 
  \end{tabular}
\end{table}

%% file: table-logmap-diso-all.tex
\begin{table}[t]
  \centering
  \begin{threeparttable}
  \caption{\footnotesize A comparison of the (partial) distributions of alignment size (measured in entity mapping counts) for two different sets of DISO ontology pairs, as generated by one ontology matching system.}
  \label{table:logmap-alignment-size-analysis}
  \begin{tabular}{c|rrr|rrr}
    \toprule  
                 & \multicolumn{3}{c}{1653 DISO ontology pairs} & \multicolumn{3}{c}{105 DISO cyber-security ontology pairs} \\
                 & \\
     Alignment   & Ontology  & Ontology & Cumulative  & Ontology  & Ontology & Cumulative  \\
     Mapping & Pair Count & Pair Count & Pair Count As & Pair Count & Pair Count & Pair Count As  \\
     Count &         & Cumulative & Proportion of   &       & Cumulative & Proportion of \\
           &         &            & Total Pairs &   &    & Total Pairs \\
    \midrule
    0 & 1006   & 1006 & 0.61 & 39 & 39 & 0.37 \\
    1 & 213  & 1219 & 0.74 & 17 & 56 & 0.53  \\
    2 & 107 & 1326 & 0.80 & 13 & 69 & 0.66  \\
    3 & 90 & 1416 & 0.86 & 11 & 80 & 0.76 \\
    4 & 48 & 1464 & 0.89 & 3 & 83 & 0.79   \\
    5 & 29  & 1493 & 0.90 & 3 & 86 & 0.82  \\
    6 & 31  & 1524 & 0.92 & 6 & 92 & 0.88  \\
    7 & 17  & 1541 & 0.93 & 2 & 94 & 0.90  \\
    8 & 14  & 1555 & 0.94 & 1 & 95 & 0.90  \\
    9 & 13  & 1568 & 0.95 & 0 & 95 & 0.90  \\
    $\geq$10 & 95 & 1653 & 1.0 & 10 & 105 & 1.0\\
    \bottomrule 
  \end{tabular}
  \end{threeparttable}
\end{table}

%% file: 3-diso-oaei.tex
\section{The DISO OAEI track}
\label{sec:diso-oaei}

\input{table-diso-oaei-merged}

The Ontology Alignment Evaluation Initiative (OAEI) \cite{OAEI2025Results} is an annual campaign for the systematic evaluation of ontology alignment systems.
The OAEI plays a key role in the benchmarking, comparison and reproducibility of ontology alignment systems.
The OAEI includes different matching tasks involving small-sized (\eg conference \cite{DBLP:journals/ws/ZamazalS17}), medium-sized (\eg anatomy \cite{anatomy2017}), and large (\eg bio-ml \cite{bio-ml2022}) ontologies.

Interoperability in the \DNS domain has been recognised as a major challenge, and a community-driven initiative such as the OAEI can contribute to the integration of the DISO ontologies. At the same time, the \DNS domain offers a promising real-world scenario and a set of non-trivial matching tasks for the campaign, as shown in this section.
The bulk ontology analysis revealed an important potential intersection among the DISO ontologies. We analysed the ontology pairs sharing a larger amount of ontology mappings (as suggested by the LogMap system) and ranked them in terms of complexity of the alignment task; that is, ontology pairs where their entities have fewer identical labels were preferred.
LogMap was selected 
to drive the task selection as it generates different types of outputs that 
enabled a fine-grained analysis.
Additional information is provided in the GUARD project report (phase 1)~\cite{city37459-diso-report-2026}.
Table \ref{table:logmap-top-8-onto-pairs} shows the top-8 candidate ontology pairs that emerged from the analytic process and form the basis for the new DISO-OAEI track.
These ontology pairs cluster into three distinct subdomains: cybersecurity (2 pairs), situation awareness (3 pairs), and smart environments (3 pairs).

\subsection{Ontology statistics and provenance}

\input{table-diso-oaei-onto-stats}

Table \ref{table:logmap-top-10-ontologies} shows the statistics of the ten selected ontologies for the DISO-OAEI matching tasks. For each ontology, we briefly describe its domain and provenance.

\smallskip
\noindent
\textit{UCO: Unified Cyber Ontology}
UCO\footnote{\url{https://unifiedcyberontology.org/}} is a domain-level ontology assigned to the DISO `cyber-security' cluster. 
It focuses on concepts such as cyber investigations, network defence, threat intelligence, malware analysis, vulnerability research, and offensive operations. UCO is an open, community-developed ontology, partly funded by the Linux Foundation, with the ambition to support the cybersecurity domain in its entirety.

\smallskip
\noindent
\textit{STIX: Structured Threat Information eXpression.}
STIX \cite{Barnum-STIX-2014} is a domain-level ontology assigned to the DISO `cyber-security' cluster. Initially developed by MITRE Corporation, STIX was designed as a language and serialisation format for exchanging cyber threat intelligence. The official STIX specification is managed by the OASIS Cyber Threat Intelligence (CTI) Technical Committee.\footnote{\url{https://oasis-open.github.io/cti-documentation/}}

\smallskip
\noindent
\textit{D3FEND.}
D3FEND \cite{Kaloroumakis-D3FEND-2021} is a domain-level ontology assigned to the DISO `cyber-security' cluster. MITRE Corporation,
its developer, refers to it as a knowledge graph of cybersecurity countermeasures, and as a framework for cybersecurity operations and strategic decision-making. Development of D3FEND was funded by the National Security Agency (NSA), the U.S. Cyber Warfare Directorate, and the U.S. Office of the Under Secretary of Defense for Research and Engineering.

\smallskip
\noindent
\textit{JC3IEDM: Joint Consultation, Command and Control Information Exchange Data Model.}
JC3IEDM \cite{NATO-JC3IEDM-2007}, standardised under NATO as STANAG 5525, is a domain-level ontology assigned to the DISO clusters `information exchange' and  `situation awareness' \cite{JC3IEDM-Valiente-2011}. The aim of the JC3IEDM is to support the exchange of consultation, command and control (C3) information, where heterogeneous command and control information systems (C2IS) are sharing military situational awareness data. %
This model has been further developed under the Multilateral Interoperability Programme (MIP) to deliver the MIP Information Model.\footnote{The MIP Information Model (2025): \url{https://www.mimworld.org/}} 

\smallskip
\noindent
\textit{mIO!: A Context Ontology for Mobile Environments.}
mIO! \cite{DBLP:conf/ekaw/Poveda-Villalon10} is a domain-level ontology assigned to DISO cluster `context awareness'.  It is a context ontology network whose aim is to model context-related knowledge that allows mobile applications to adapt their behaviour based on user context. 
mIO! was developed by the Ontology Engineering Group at University of Madrid.\footnote{\url{https://oeg.fi.upm.es/index.php/en/ontologies/index.html}}

\smallskip
\noindent
\textit{Brick: A Uniform Metadata Schema for Buildings.}
Brick\footnote{\url{https://brickschema.org/}} is a domain-level ontology assigned to DISO cluster `smart buildings' (`smart environments'). It is an open-source project whose aim is to standardise descriptions of physical, logical and virtual aspects of buildings, and the relationships between them. The Brick consortium comprises commercial and academic members.

\smallskip
\noindent
\textit{Facility.}
Facility is a CCO (Common Core Ontology)\footnote{\url{https://www.commoncoreontologies.org}} mid-level ontology assigned to DISO cluster `cco-modules', within cluster `mid-level'.  This mid-level ontology was designed to represent facilities (such as buildings, campuses, etc.) that serve some specific purpose and which are common in multiple domains.

\smallskip
\noindent
\textit{ThinkHome: Smart Home Ontology for Human Activity Recognition.}
ThinkHome \cite{DBLP:journals/ejes/ReinischKIK11} is a domain-level ontology assigned to DISO cluster `smart homes' (`smart environments'). It is an ontology intended to support smart systems, especially those with a focus on home control, and particularly on energy efficiency. It consists of five component ontologies that address individual aspects of the target application domain: actors, buildings, energy resources, processes and weather. ThinkHome was developed at Technical University Vienna.\footnote{\url{https://www.auto.tuwien.ac.at/index.php/projectsites/155-thinkhome}}

\smallskip
\noindent
\textit{SmartEnv: Smart Home Environments.}
SmartEnv \cite{DBLP:journals/semweb/AlirezaieHB18,DBLP:journals/sensors/KockemannARTAML20} is a domain-level ontology assigned to DISO cluster `smart homes' (`smart environments'). SmartEnv, initially referred to as `E-care@home' in the literature \cite{DBLP:journals/sensors/AlirezaieRKKKBT17}, is a network of 8 component ontologies modelling various aspects of smart homes: temporal, spatial, objects, events, agents, observations, etc. Its notion of `smart' refers to environments (such as homes) that are monitored by sensors of various kinds.

\smallskip
\noindent
\textit{CityOWL.}
CityOWL is a domain-level ontology assigned to DISO cluster `smart cities' (`smart environments').  It is an OWL rendering of a standard schema defined by the Open Geospatial Consortium called CityGML\footnote{\url{https://www.ogc.org/standards/citygml/}}. The standard defines a conceptual model for representing virtual 3D models (aka digital twins) of cities. The CityOWL ontology obtained for DISO was acquired from a URL associated with the Computer Science Laboratory for Image and Information Systems at the University of Lyon.\footnote{\url{https://liris.cnrs.fr/}}

\subsection{Alignment evaluation}

We executed several state-of-the-art systems on the DISO-OAEI matching tasks with two main objectives: \textit{(i)} to assess the complexity of the tasks, and \textit{(ii)} to generate silver-standard alignments for the track. We selected the following systems: AML~\cite{aml-2025}, BertMap~\cite{bertmap2022} (with variations using different BERT-based pre-trained models), BertMapLt, LogMap \cite{logmap2011}, LogMapLt, LogMapLLM \cite{logmapllm-eacl-2026}, Matcha \cite{matcha2023}, ALOD2Vec~\cite{DBLP:conf/semweb/PortischP21a}, ATMatcher \cite{DBLP:conf/semweb/HertlingP22}, Fine-TOM \cite{DBLP:conf/semweb/KnorrP21} and KGMatcher \cite{DBLP:conf/semweb/FallatahZH21a}. Table \ref{tab:eval} presents the mappings computed by the ontology alignment systems in each of the DISO-OAEI tasks. Note that LogMapLLM, ALOD2Vec, ATMatcher, Fine-TOM, and KGMatcher were not able to produce mappings on all matching tasks. It can also be observed that the systems generated highly disparate sets of mappings in the DISO-OAEI tasks. For example, in the \textit{JC3IEDM-mIO!} task, LogMap produces 234 mappings while AML generates 34.

\input{table-diso-mappings-consensus}

\paragraph{Consensus alignments.}
We have computed consensus alignments of vote 2 and 3 (\ie mappings suggested by at least two or three systems, respectively). Note that, when there are several systems of the same family (\ie systems participating with several variants), their (voted) mappings are only counted once in order to reduce bias.
Table~\ref{tab:eval} shows
the number of consensus mappings voted by 2, 3 and all systems (Con-2, Con-3 and Con-all, respectively) and the number of system families contributing to the consensus~(\#SF). For example, in the \textit{JC3IEDM-mIO!} task, 77 mappings were suggested by at least 2 system families, while only 11 mappings were recommended by all 8  families.

\input{table-manual-analysis}

\paragraph{Manual assessment.} Consensus alignment can serve as a reference for the agreement across systems; they are, however, not complete and may also include errors, as it only requires two systems to agree on an incorrect mapping \cite{oaei-phenotype}. Hence, we have performed a manual revision of the Con-2 consensus mappings and the mappings uniquely generated by a system or system family. Unique mappings may include potentially valid correspondences that were not identified by any other system family. Table \ref{tab:manual} presents the results of the manual validation. Consensus mappings are typically precise, with 94\% of the mappings in \textit{JC3IEDM-Facility} classified as correct. Con-2 was less accurate for the \textit{JC3IEDM-Brick} task. The quality of unique mappings varied across systems and tasks.
For example, 80\% and 13\% of the unique mappings identified by LogMap and AML in \textit{ThinkHome-Brick}, respectively, were annotated as correct. The manual revision led to the creation of a silver-standard for the DISO-OAEI tracks by merging the correct unique mappings and the correct Con-2 mappings.
The manual validation was performed primarily by the authors.
However, a broader validation involving multiple domain experts, together with an inter-annotator agreement analysis, is planned in the context of the OAEI 2026 campaign. The quality of the silver reference will also improve as more systems contribute mappings, and it will be updated with results from the OAEI participants.

\begin{figure}[t!]
    \centering    
    \includegraphics[width=1.0\textwidth]{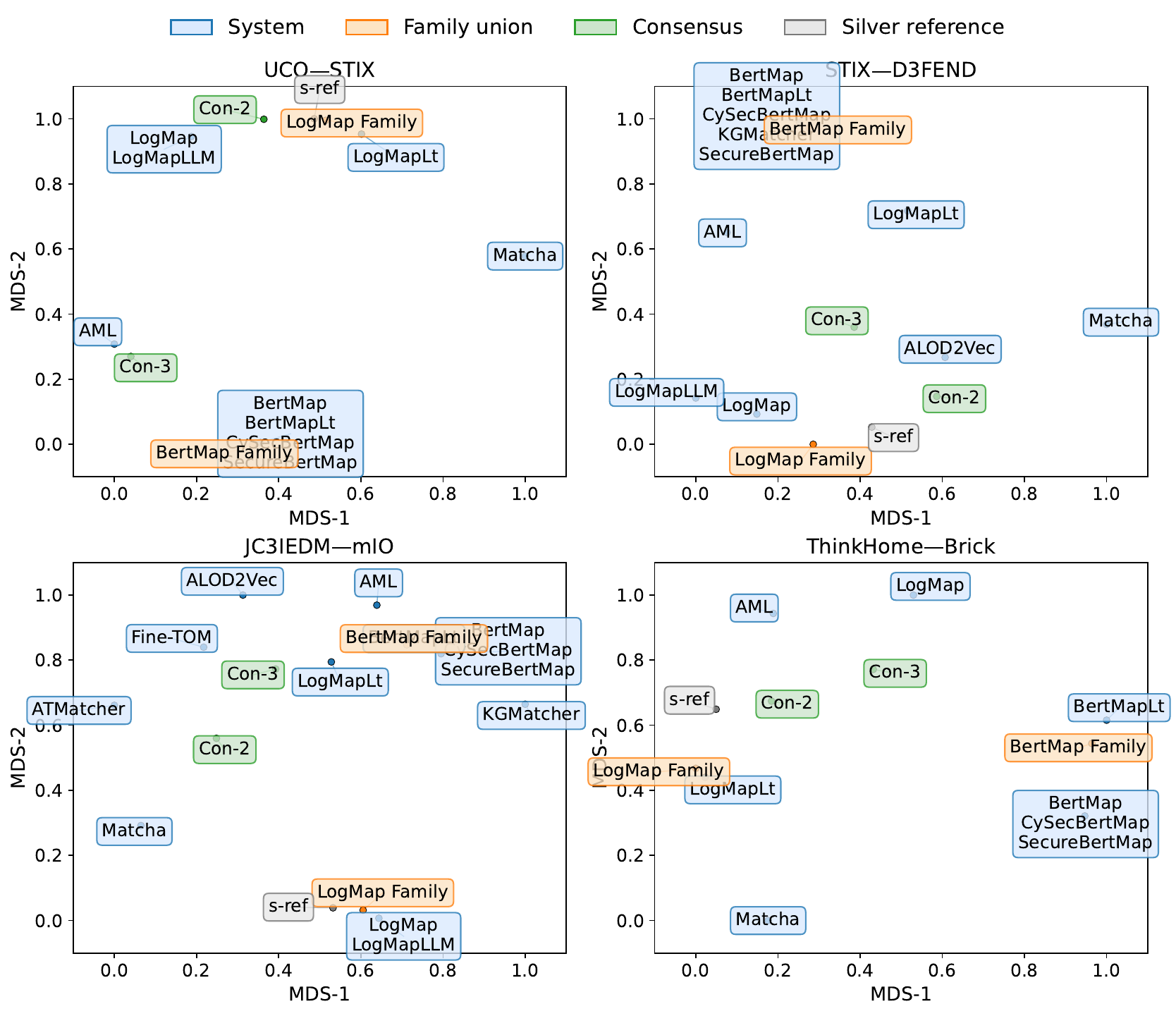}
      \caption{Visual representation of the Jaccard distances among mapping sets.}
    \label{fig:distances}
\end{figure}

\paragraph{Mapping comparison.}
Table \ref{tab:eval} and the consensus mapping extraction have revealed a significant disparity among the evaluated systems, which is further illustrated in Figure~\ref{fig:distances}. This confirms that the selected matching task involves the discovery of non-trivial mappings.
The figure presents two-dimensional scatterplots of the Jaccard distances between the compared mapping sets. Proximity to the consensus mapping set is not necessarily an indicator of better quality; rather, it reflects a higher agreement with the consensus. For example, in \textit{JC3IEDM-mIO!}, LogMapLt is close to the Con-3 consensus but far from \textit{s-ref} (the manually-curated silver-standard reference). System families are also visualised, representing the aggregation of all mappings produced by the corresponding system variants.
We emphasise that the silver-standard is derived from the outputs of the participating systems, its recall is inherently bounded by what these systems can collectively discover, and it will favour systems that resemble those used in its construction. Thus, it should be regarded as a partial reference rather than a complete gold standard. 
For the same reason, we prefer not to report the actual precision and recall values of the participating systems with respect to the silver standard, but rather the proximity among the relevant mapping sets, as shown in Figure~\ref{fig:distances}.

\paragraph{Ontology compatibility.} Reasoning with the input ontologies and the mappings may lead to logical errors (\eg unsatisfiabilities).  An alignment $\M$ is incoherent if $\On_1 \cup \On_2 \cup \M \models A \sqsubset \bot$ (for any class A).
Table \ref{tab:unsat} shows the logical errors led by system-computed mappings and the generated consensus and silver-standard alignments. It is worth noting that even systems implementing (approximate) reasoning techniques like AML and LogMap \cite{kais_sjg_2016} lead to a large number of unsatisfiabilities (\eg \textit{UCO-STIX} and \textit{STIX-D3FEND}). This highlights a modelling disagreement among the DISO-OAEI ontologies, bringing an interesting challenge to the OAEI from the logical and reasoning point of view.
The silver-standard alignment, which involved manual curation, also leads to unsatisfiable classes, emphasising the need for reasoning techniques within the ontology alignment pipeline.
For the DISO-OAEI 2026 track, the silver standard has been treated to fix logical errors and annotate incoherent mappings.

\input{table-logical-errors3}

\begin{figure}[t!]
    \centering    
    \includegraphics[width=0.80\textwidth]{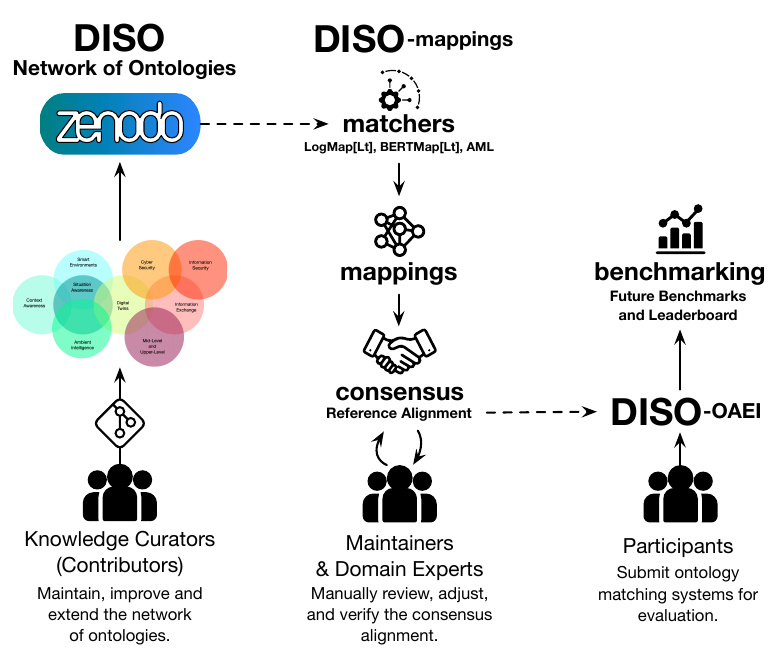}
      \caption{The DISO ecosystem. The \textit{diso} ontology collection aims to be extended through a collective effort by contributors and released through Zenodo. The diso-mapping pipeline runs a configurable, extendable set of matchers and aggregates their outputs into a family-based consensus alignment. The alignment is reviewed by maintainers and domain experts, and serves as a partial silver-standard reference alignment, forming the foundation for the \textit{diso-oaei} track, where OAEI systems will be benchmarked.}
    \label{fig:ecosystem}
\end{figure}

\section{The DISO ecosystem}
\label{sec:ecosystem}

Resources are made available through a small ecosystem of three repositories, as depicted in Figure \ref{fig:ecosystem}. 
In particular, the collection of \DNS ontologies is curated in the \textit{diso} repository\footnote{\url{https://github.com/city-artificial-intelligence/diso}} and mirrored to Zenodo\footnote{\url{https://doi.org/10.5281/zenodo.20059506}} upon every major release. Since some of the ontologies are packaged as networked component ontologies, we ship both a canonical distribution and a compact distribution. The compact distribution merges networked ontologies into a single file for the convenience of downstream consumers. For instance, the compact distribution is consumed by our alignment extraction pipeline \textit{diso-mappings}\footnote{\url{https://github.com/city-artificial-intelligence/diso-mappings}}, which is the second repository in our ecosystem. It allows contributors to easily download the compact distribution, register new matching systems, and iteratively improve the reference alignment.
\textit{diso-mappings} is a Python pipeline that consumes the DISO compact distribution and produces pairwise alignments between selected ontology pairs. It uses a configurable set of ontology matching systems (AML, LogMap, LogMapLt, BERTMap, and BERTMapLt by default; with an extendable Matcher base class that enables custom external matchers). The pipeline aggregates the per-system alignments into a 
consensus alignment via a family-based voting mechanism. This alignment is then manually verified, adapted and used as the basis for a partial silver-standard reference alignment. The computed alignments can easily be reproduced by following the steps within the repository’s make workflow.

\textit{diso-oaei}\footnote{\url{https://github.com/city-artificial-intelligence/diso-oaei}} is the home for the new DISO-OAEI track introduced in the OAEI 2026 edition. It contains the datasets, evaluation metrics, participation instructions, and will also host the OAEI benchmarking results.

%% file: table-diso-oaei-merged.tex
\begin{table}[t]
  \centering
  \scriptsize
  \begin{threeparttable}
  \caption{\footnotesize The top-8 ontology pairs that emerged from the LogMap-driven analytic process. `SA' refers to `situation awareness'; `CA' refers to `context awareness'.}
  \label{table:logmap-top-8-onto-pairs}
  \begin{tabular}{l|cc|cc|rrrrr}
    \toprule
    \multirow{2}{*}{Task Subdomain~~~} &
    \multicolumn{2}{c|}{Source Ontology} & \multicolumn{2}{c|}{Target Ontology} &
    \multicolumn{5}{c}{LogMap Alignment} \\\cmidrule{2-10}
     & Name &  Cluster & Name & Cluster  & ~~~All & ~Cls & Oprop & Dprop & Ind \\
    \midrule
    \multirow{2}{*}{cybersecurity} &
    UCO &  cybersecurity & STIX & cybersecurity & 131 & 43 & 4 & 84 & 0  \\
    & STIX & cybersecurity & D3FEND & cybersecurity & 37 & 31 & 0 & 6 & 0 \\
    \addlinespace
    \multirow{3}{*}{SA} & JC3IEDM & SA & mIO! &CA & 219 & 12 & 0 & 0 & ~~207\\
    & JC3IEDM & SA & Brick & smart buildings & 49 & 6 & 0 & 0 & 43 \\
    & JC3IEDM & SA & Facility & cco-modules & 44 & 44 & 0 & 0 & 0 \\
    \addlinespace
    \multirow{3}{*}{smart envir} &
    ThinkHome & smart homes & Brick & smart buildings & 33 & 24 & 0 & 1 & 7 \\
    & Brick & smart buildings & SmartEnv & smart homes & 32 & 25 & 0 & 0 & 7 \\
    & CityOWL & smart cities & Brick  & smart buildings & 24 & 16 & 1 & 0 & 7 \\
    \bottomrule 
  \end{tabular}
  \end{threeparttable}
\end{table}

%% file: table-diso-oaei-onto-stats.tex
\begin{table}[t]
  \centering
  \scriptsize
  \begin{threeparttable}
  \caption{\footnotesize Quantitative descriptions of the 10 ontologies appearing in the DISO-OAEI matching tasks. The metrics are as reported by the Prot\'eg\'e  (\url{https://protege.stanford.edu/}).}
  \label{table:logmap-top-10-ontologies}
  \begin{tabular}{lrrrrrrrrrr}
    \toprule 
     Metric & UCO  & STIX  & D3FEND & JC3IEDM & mIO!  & Brick & Facility & ThinkHome & SmartEnv & CityOWL \\
    \midrule
    Axioms &  11,552 & 3,903 & 32,832 & 26,622 & 7,727 & 117,811 & 7,083 & 8,122 & 4,146 & 7,387 \\
    Class count & 429 & 88 & 3,495 & 2,922 & 624 & 2,305 & 859 & 1,109 & 161 & 516 \\
    Object property count & 192 & 68 & 205 & 619 & 364 & 131 & 146 & 406 & 242 & 522 \\
    Data property count & 581 & 386 & 42 & 313 & 310 & 80 & 12 & 303 & 50 & 162 \\
    Individual count & 507 & 4 & 2,340 & 4,088 & 520 & 8,556 & 81 & 54 & 17 & 17 \\ 
    Annotation property count & 43 & 6 & 35 & 2 & 24 & 165 & 40 & 22 & 44 & 42 \\
    \addlinespace
    \textbf{Class axioms} & \\
    SubClassOf & 444 & 1270 & 5,391 & 2,567 & 627 & 2,494 & 925 & 2,344 & 401 & 1,250 \\
    EquivalentClasses & 0 & 0 & 0 & 272 & 36 & 105 & 25 & 282 & 6 & 2 \\
    DisjointClasses & 8 & 18 & 8 & 0 & 62 & 15 & 23 & 123 & 21 & 36 \\
    \addlinespace
    \textbf{Object property axioms} & \\
    SubObjectPropertyOf & 11 & 0 & 250 & 0 & 92 & 19 & 84 & 61 & 164 & 48 \\
    EquivalentObjectProperties & 0 & 0 & 0 & 0 & 0 & 4 & 0 & 0 & 13 & 13 \\
    InverseObjectProperties & 9 & 0 & 41 & 116 & 70 & 9 & 63 & 48 & 80 & 7 \\
    ObjectPropertyDomain & 10 & 89 & 2 & 503 & 344 & 0 & 127 & 350 & 177 & 517 \\
    ObjectPropertyRange & 181 & 71 & 4 & 503 & 337 & 2 & 128 & 340 & 173 & 511 \\ 
    \addlinespace
    \textbf{Data property axioms} & \\    
    SubDataPropertyOf & 40 & 1 & 42 & 0 & 92 & 6 & 0 & 12 & 13 & 9 \\
    EquivalentDataProperties & 0 & 0 & 0 & 0 & 0 & 0 & 0 & 0 & 0 & 0 \\
    DataPropertyDomain & 4 & 448 & 2 & 313 & 304 & 2 & 8 & 82 & 48 & 162 \\
    DataPropertyRange & 579 & 385 & 17 & 313 & 284 & 2 & 8 & 277 & 43 & 159 \\
    \addlinespace
    \textbf{Individual axioms} & \\    
    ClassAssertion & 558 & 0 & 1,194 & 4,088 & 904 & 18,882 & 83 & 53 & 29 & 25 \\
    \bottomrule 
  \end{tabular}
  \end{threeparttable}
\end{table}

%% file: table-diso-mappings-consensus.tex
\setlength{\tabcolsep}{2.51pt}
\begin{table}[t!]
    \centering
    {\scriptsize
    \caption{Number of system and consensus mappings for the selected matching tasks. \#SF: number of system families contributing to the consensus.  Con-x: consensus mappings with `x' votes. Con-all: consensus mapping supported by all system families.}
    \label{tab:eval}
    \vspace{0.1cm}
    \begin{tabular}{|l||c|c|c|c|c|c||c|c|c|c|}
    \hline 
        \multirow{2}{*}{\textbf{Matching task}} & \multicolumn{6}{c||}{\textbf{System mappings}} & \multicolumn{4}{c|}{\textbf{Consensus mappings}} \\\cline{2-11} 
        & \textbf{AML} 
        & \textbf{BertMap} & \textbf{BertMapLt}
        &
        \textbf{LogMap} & 
        \textbf{LogMapLt}  & 
         \textbf{Matcha} & 
               
        \textbf{\#SF} &
        
        \textbf{Con-2} & \textbf{Con-3} & \textbf{Con-all} 
        \\\hline\hline 

        \textbf{UCO-STIX} & 43 & 21 & 21  & 131 & 177 & 692 & 4 & 171 & 39 & 21   \\\hline

        \textbf{STIX-D3FEND} & 27 & 17 & 18 & 37 & 28  & 102  & 6 & 45  & 32 & 17    \\\hline

        \textbf{JC3IEDM-mIO!} & 34 & 22  & 26 & 234 & 37 & 167 & 8 & 77 & 46 & 11     \\\hline

        \textbf{JC3IEDM-Brick} & 45 &  23 & 23 & 57 & 105 & 776 & 4 & 108 & 27  & 22   \\\hline

        \textbf{JC3IEDM-Facility} & 43 & 28 & 29 & 44 & 37  & 43  & 8 & 71  &  41 & 2  \\\hline

        \textbf{ThinkHome-Brick} & 112 & 13 & 17 & 55 & 141 & 312  & 4 & 95  & 58 & 16   \\\hline

        \textbf{Brick-SmartEnv} & 32 & 12  & 22 & 39 & 52 & 322  & 4 & 48  & 28 & 22    \\\hline

        \textbf{CityOWL-Brick} & 38 & 15  & 22 & 27 & 58 &  402 & 4 & 64  & 28 & 23 \\\hline
        

    \end{tabular}
    } 
\end{table}

%% file: table-manual-analysis.tex









\setlength{\tabcolsep}{3.5pt}
\begin{table}[t]
    \centering
    {\scriptsize
    \caption{Manual validation of unique and consensus mappings.
    For each system family, we provide the number of unique mappings and the ratio of correct ones in brackets.
    For Matcha, the ratio represents an approximation given the large number of generated mappings.
    \textit{Others} aggregates ALOD2VEc, ATMatcher, Fine-TOM and KGMatcher.}.
    \label{tab:manual}
    \vspace{0.2cm}
    \begin{tabular}{|l||c|c|c|c|c||c|c||c|}
    \hline
        \multirow{2}{*}{\textbf{Matching Task}}
            & \multicolumn{5}{c||}{\textbf{Unique system mappings by family}}
            & \multicolumn{2}{c||}{\textbf{Consensus-2 mappings}}
            & \textbf{Silver} \\
        \cline{2-9}
            & \textbf{AML}
            & \textbf{BertMap}
            & \textbf{LogMap}
            & \textbf{Matcha}
            & \textbf{Others}
            & \textbf{\# Total}
            & \textbf{\% Correct}
            & \textbf{\# Total} \\
        \hline\hline
        \textbf{UCO-STIX}         & 1 (0\%)     & 1 (0\%)    & 43  (84\%) & 516 (10\%) & 0  (0\%)    & 171 & 91\% & 194  \\\hline
        \textbf{STIX-D3FEND}      & 2 (0\%)     & 0 (0\%)    & 10  (20\%) & 63  (10\%) & 0  (0\%)    & 45  & 76\% & 39   \\\hline
        \textbf{JC3IEDM-mIO!}     & 3 (0\%)     & 0 (0\%)    & 199 (97\%) & 113 (7\%)  & 65 (9\%)  & 77  & 87\% & 274  \\\hline
        \textbf{JC3IEDM-Brick}    & 3 (0\%)     & 2 (50\%) & 38  (5\%)  & 662 (2\%)  & 0  (0\%)    & 108 & 67\% & 75   \\\hline
        \textbf{JC3IEDM-Facility} & 6 (0\%)     & 3 (33\%) & 5 (60\%) & 0   (0\%)  & 34 (0\%)    & 44  & 94\% & 44  \\\hline
        \textbf{ThinkHome-Brick}  & 45 (13\%) & 0 (0\%)    & 80  (47\%) & 299 (6\%)  & 0  (0\%)    & 95  & 92\% & 134  \\\hline
        \textbf{Brick-SmartEnv}   & 3 (0\%)     & 0 (0\%)    & 36  (31\%) & 274 (10\%) & 0  (0\%)    & 48  & 77\% & 53   \\\hline
        \textbf{CityOWL-Brick}    & 6 (0\%)     & 0 (0\%)    & 21  (57\%) & 339 (5\%)  & 0  (0\%)    & 64  & 83\% & 68   \\\hline
    \end{tabular}
    }
\end{table}


%% file: table-logical-errors3.tex


 \setlength{\tabcolsep}{0.5pt}
 \begin{table}[t!]
     \centering
     {\scriptsize
         \caption{Unsatisfiable classes led by system, consensus and silver-standard mappings, computed over the whole merged ontology
         with 
         the OWL 2 El reasoner ELK \cite{elk2014}, marked by $\geq$ (lower-bound), or a complete OWL 2 DL reasoner (JFact \cite{factpp2006}, otherwise HermiT \cite{hermit2014}).
         \textit{incons.} denotes a merged TBox detected as inconsistent by ELK.}
     \label{tab:unsat}
     \vspace{0.1cm}
     \begin{tabular}{|l||c|c|c|c|c|c||c|c|}
     \hline
         \multirow{2}{*}{\textbf{Matching task}} & \multicolumn{6}{c||}{\textbf{System mappings}} & \multicolumn{2}{c|}{\textbf{Aggregated mappings}} \\\cline{2-9}
        & \textbf{AML}
         & \textbf{BertMap} & \textbf{BertMapLt}
         &
         \textbf{LogMap} &
         \textbf{LogMapLt}  &
          \textbf{Matcha} &

         \textbf{Con-2} &  \textbf{Silver}
         \\\hline\hline

         \textbf{UCO-STIX} & 122 (24\%)  & 110 (21\%)   & 110 (21\%)  & 134 (26\%) & 131 (25\%) &  127 (24\%) & 122 (24\%) & 133 (25\%)
         \\\hline

         \textbf{STIX-D3FEND} &  2448 (68\%) & 2441 (68\%) & 2444 (68\%) &  2516 (70\%) &  2516 (70\%) & 2509 (70\%) & 2509 (70\%) & 2506 (70\%)\\\hline

         \textbf{JC3IEDM-mIO!} & 0 & 0 & 0 & 0 & 0 & 0& 0 &0 \\\hline

         \textbf{JC3IEDM-Brick} & 0 & 0 & 0 & 0 & 0 & 0 & 0 & 0\\\hline

         \textbf{JC3IEDM-Facility} & 0 & 0 & 0 & 0 & 0 & 0 & 0 & 0\\\hline

         \textbf{ThinkHome-Brick} & 1123 (33\%) & 0  & $\geq$49 (1.4\%)  & 0  & 1375 (40\%) & $\geq$652 (19\%) & 907 (27\%) & 874 (26\%) \\\hline

         \textbf{Brick-SmartEnv} & $\geq$12 & 0 & $\geq$11 & 0 & $\geq$844 (34\%) & incons. & $\geq$844 (34\%) & $\geq$842 (34\%)\\\hline

         \textbf{CityOWL-Brick} & 26 (0.9\%) & 0 & 20 (0.7\%) & 0  & 20 (0.7\%) & 279 (9.9\%) & 30 (1.1\%) & 21 (0.7\%)\\\hline

     \end{tabular}
     }
 \end{table}


%% file: 4-conclusions.tex
\section{Conclusions and sustainability of the resources}
\label{sec:conclusions}

In this work, we have curated and described a collection of publicly available OWL ontologies covering the \DNS domain, and proposed a new OAEI track for evaluating ontology matching systems on this collection. 
Our analysis identified eight candidate ontology pairs, spanning three distinct subdomains, supporting non-trivial ontology matching tasks across classes, properties, and instances. 
We released three resources in support of a rich and diverse track: \textit{diso}, the curated ontology collection for defence, intelligence and national security; \textit{diso-mappings}, a reproducible pipeline that aggregates the output of several state-of-the-art matchers into a consensus alignment via family-based voting; and \textit{diso-oaei}, the home page for the new OAEI 2026 track.
The consensus alignment was manually reviewed \textit{(by the authors)}, leading to a silver-standard reference alignment.

This effort has been conducted as part of the GUARD project, funded by  The Turing Defence \& Security Grand Challenge.
We are actively collaborating with experts from the UK Government intelligence community, the NDTP (National Digital Twin Programme), and the Dstl (Defence Science and Technology Laboratory), who will support the development and maintenance of the DISO ecosystem. The DISO ecosystems aim at serving both the broader research community and the UK's national needs on \DNS with regard to the improvement of the interoperability of the relevant domain ontologies.

The \DNS domain presents distinctive challenges for ontology matching: source ontologies span heterogeneous granularities (\eg upper-level, mid-level, temporal), numerous subdomains, and gold standard ground-truth coverage is limited. 
The OAEI community-driven evaluation will help to inform the future direction of research in this domain. The OAEI evaluation and the collaboration with domain experts will also support the extensions and improvement of the DISO ecosystem: ontologies, matching tasks, silver-standards, and, ultimately, the creation of an integrated network of DISO ontologies.
For the OAEI 2026, we also plan to include automated benchmarking procedures, an active leaderboard, and a candidate mapping ranking task as in the bio-ml track \cite{bio-ml2022} to attract the participation of machine learning systems.

DISO is not the first initiative to gather domain-specific ontologies. A prominent example is BioPortal \cite{bioportal2009}, which has become the reference repository for biomedical ontologies. Similarly, the National Center for Ontological Research (NCOR) is developing a National Security Ontology Foundry,\footnote{\url{https://ncor-network.org/}} with which we intend to explore potential collaborations. As DISO evolves, we will also explore adopting the OntoPortal technology to host and manage its ontologies~\cite{DBLP:conf/semweb/JonquetGBDFKRRSVM23}.\footnote{\url{https://ontoportal.org/}}